\documentclass{article}

\usepackage{microtype}
\usepackage{graphicx}
\usepackage{subfigure}
\usepackage{booktabs} 
\usepackage{multirow}
\usepackage{comment}
\usepackage{natbib}
\usepackage{placeins}
\usepackage{hyperref}

\usepackage[accepted]{icml2023}

\usepackage{amsmath}
\usepackage{amssymb}
\usepackage{mathtools}
\usepackage{amsthm}

\usepackage[capitalize,noabbrev]{cleveref}

\theoremstyle{plain}

\theoremstyle{definition}

\theoremstyle{remark}

\usepackage[textsize=tiny]{todonotes}

\icmltitlerunning{Inferring Hierarchical Structure in Multi-Room Maze Environments}

\begin{document}

\twocolumn[
\icmltitle{Inferring Hierarchical Structure \\ in Multi-Room Maze Environments}



\icmlsetsymbol{equal}{*}

\begin{icmlauthorlist}
\icmlauthor{Daria de Tinguy}{yyy}
\icmlauthor{Toon Van de Maele}{yyy}
\icmlauthor{Tim Verbelen}{comp}
\icmlauthor{Bart Dhoedt}{yyy}

\end{icmlauthorlist}

\icmlaffiliation{yyy}{Department of Information Technology, University of Ghent, Ghent, Belgium}
\icmlaffiliation{comp}{Verses AI, Vancouver, Canada}

\icmlcorrespondingauthor{Daria de Tinguy}{Daria.detinguy at ugent.be}

\icmlkeywords{Machine Learning, ICML}

\vskip 0.3in
]



\printAffiliationsAndNotice{\icmlEqualContribution} 

\begin{abstract}

Cognitive maps play a crucial role in facilitating flexible behaviour by representing spatial and conceptual relationships within an environment. The ability to learn and infer the underlying structure of the environment is crucial for effective exploration and navigation. This paper introduces a hierarchical active inference model addressing the challenge of inferring structure in the world from pixel-based observations. We propose a three-layer hierarchical model consisting of a cognitive map, an allocentric, and an egocentric world model, combining curiosity-driven exploration with goal-oriented behaviour at the different levels of reasoning from context to place to motion. This allows for efficient exploration and goal-directed search in room-structured mini-grid environments. 

\end{abstract}

\section{Introduction}



Understanding and navigating through complex environments is a fundamental capability for intelligent agents. Traditional approaches rely on metrical or topological maps to guide agent movement. However, in order for an agent to be truly autonomous, it must possess the ability to learn and adapt to its surroundings.
In order to understand a complex and possibly aliased world and navigate in it, both spatial hierarchy, i.e. capturing spatial structures and relationships, and temporal hierarchy are required to plan long-term navigation schemes.

Several approaches have been proposed to learn the structure of the world in the context of navigation. \citet{CSCG} proposes a clone structured graph representation of the environment to disambiguate aliased observations. \citet{weird_HAIF} presents a deep hierarchical model based on active inference and casts structure learning as a Bayesian model reduction problem. \citet{pezzulo_Hgenerative_model} introduces a hierarchical generative model learning and recognising maze structures based on specific localisation in a global prefixed frame. 
While all these generative models aim to capture the underlying structure and dynamics of the world, these are typically limited to small simulations with discrete state and observation spaces. 

Addressing this aspect, recent approaches like G-SLAM \cite{g-slam} and Dreamer \cite{dreamer} use deep neural networks to learn generative world models from high-dimensional observations such as pixels. However, as these capture the world in a flat latent state space, these models struggle with long-term planning, especially in aliased environments.

\begin{figure}[t!]
\vskip 0.05in
\begin{center}
\centerline{\includegraphics[width=5cm]{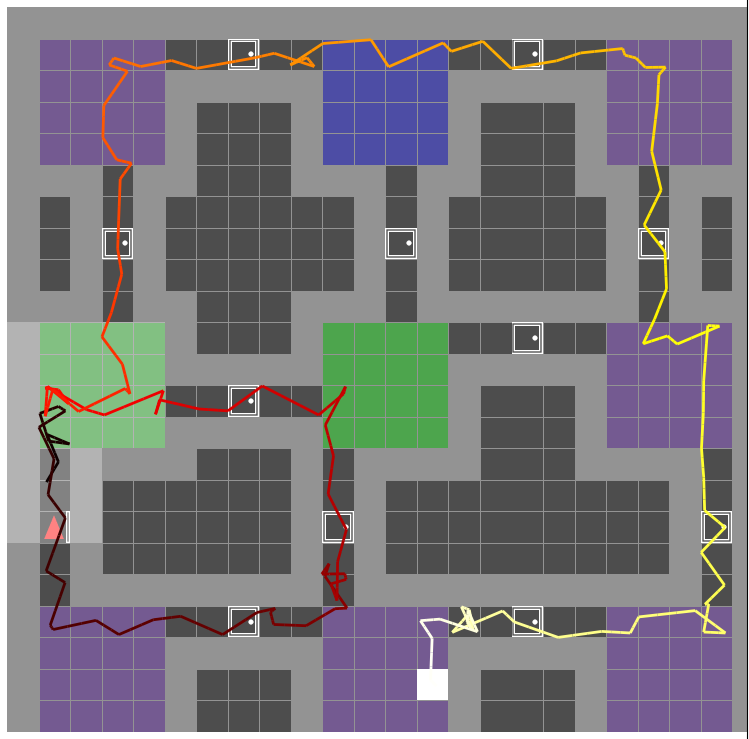}}
\caption{Example of a 3 $\times$ 3 rooms mini-grid environment and our model navigation in it during an exploration and goal-reaching task, where the starting position is the red triangle. Noise on the visualised path was added in post-processing for observing superposed visits on a single tile.}
\label{img:3x3_env}
\end{center}
\vspace{-8mm}
\end{figure}

\begin{figure*}[t!]
\begin{center}
  \includegraphics[width=14cm]{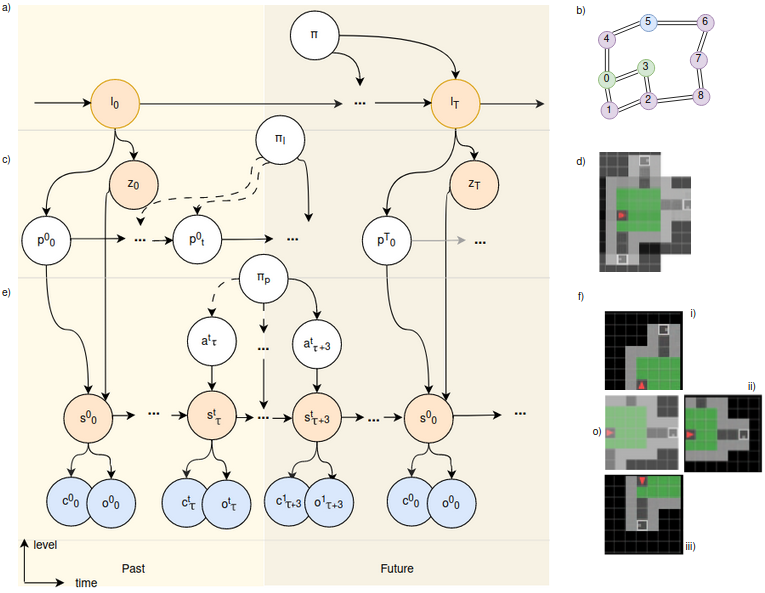}
  \caption{The left shows the graphical model of the 3-layer hierarchical Active inference model consisting of a) the cognitive map, b) the allocentric model, and c) the egocentric model, each operating at a different time scale. The orange circles represent latent states that have to be inferred, the blue circles denote observable outcomes and the white circles are random variables to be inferred by planning. The right part visualises the representation at each layer. The cognitive map is represented as d) a topological graph composed of all the locations ($l$) and their connections, in which each location is stored in a distinct node. The allocentric model e) infers place representations ($z$) by integrating sequences of observations ($s$) and poses ($p$), from which the room structure can be generated. The egocentric model f) imagines future observations given the current position, state ($s$), and possible actions ($a$). Here o) depicts an actual observation ($o$) and the predicted observations of the possible actions left i), forward ii), and right iii).}
  \label{img:HAIF_more}
\end{center}
\vspace{-4mm}
\end{figure*}

This paper proposes a pixel-based hierarchical model exhibiting both spatial and temporal hierarchies. The model is geared towards learning the structure of maze mini-grid environments \cite{gym_minigrid}.
A maze consists of interconnected, visually similar rooms with variations in shape, size, and colour as depicted in Fig \ref{img:3x3_env}. Within the maze, there is one white goal tile, which the agent must find.
Our model consists of amortised inference models at the lower levels, which are trained on pixel data, for representing movement and pose in egocentric resp. allocentric reference frames, combined with a graph-structured model at the top to capture the maze structure. 
The model navigation is based on a principled active inference approach, which balances goal-directed behaviour and epistemic foraging through information gain~\cite{AIF_learning}. Moreover, the planning happens at different temporal time scales at each level in the hierarchy, allowing for long-term decision-making.

\section{Method}


The active inference framework~\cite{parr_active_2022} is built on the premise that intelligent agents minimise their surprise. An active inference agent entails an internal generative model aiming to best explain the causes of external observation and the consequences of its actions through the minimisation of surprise or prediction error, which is also known as free energy (FE). Agents minimise this quantity with respect to model parameters in learning and with respect to action in planning~\cite{AIF_learning, nav_aif}.

We propose a hierarchical generative model consisting of three layers functioning at nested timescales (see Fig~\ref{img:HAIF_more}). From top to bottom: the cognitive map, creating a coherent topological map, the allocentric model, representing space, and the egocentric model, managing motions. The structure of the environment is inferred over time by agglomerating visual observations into representations of distinct places (e.g. rooms) at higher levels discovering the connectivity structure of the maze as a graph.

\textbf{The cognitive map}: The top layer in the generative model, illustrated in Fig~\ref{img:HAIF_more}a  functions at the coarsest time scale ($T$), each tick at this time scale corresponds to a distinct location ($l_T$) integrating the initial positions ($p^T_0$) of the place ($z^T$). These locations are depicted as nodes in a topological graph, as shown in Fig~\ref{img:HAIF_more}d. Edges between nodes are added as the agent moves from one location to another, effectively learning the maze structure. In order to maintain the spatial structure between locations, the agent keeps track of its relative rotation and translation using a continuous attractor network (CAN) as in ~\cite{ratslam_hippo}. Hence the cognitive map forms a comprehensive representation of the environment, enabling the agent to navigate by formulating believes over its surroundings. 


\textbf{The allocentric model}: The middle layer, illustrated in Fig~\ref{img:HAIF_more}b, plays a crucial role in constructing a coherent understanding of the environment, denoted as $z_T$. This model functions at a finer time scale ($t$), forming a belief over the place by integrating a sequence of observations ($s^T_t$) and poses ($p^T_t$) to generate this representation \cite{GQN_origin, GQN_Toon}. Fig \ref{img:HAIF_more}e and Fig \ref{img:room_generation} showcase the resulting place defining the environment given accumulated observations. As the agent moves from one place to another, once the current observations do not align with the previously formed prediction about the place, the allocentric model resets its place description and gathers new evidence to construct a representation of the newly discovered room ($z_{T+1}$), advancing by one tick on the coarser time scale and resetting the mid-level time scale $t$ to 0.

\textbf{The egocentric model}: The lowest layer, illustrated in Fig~\ref{img:HAIF_more}c, has the finest time scale ($\tau$). To evolve in time this model requires the prior state ($s^t_{\tau}$) and current action ($a^t_{\tau+1}$) to infer the current observation $(o^t_{\tau+1}$) \cite{generative_models_oz}. Based on its current position, the model generates possible future trajectories while considering the constraints imposed by the environment, such as the inability to pass through walls (achieved by discerning the cause-effect relationship between actions and observations). Fig \ref{img:HAIF_more}f illustrates the current observation in the middle o) and shows the imagined potential observations if the agent were to turn left i), right iii), or move forward ii).

The model operates within a hierarchical active inference scheme, planning at different time scales.
The cognitive map plays a vital role in long-term navigation by handling place connectivity, allowing the model to plan the locations to visit ($\pi$) at a high level, the poses to visit within each place ($\pi_l$) at a mid-level, and determining the best action policy ($\pi_p$) at the low level while considering obstacles such as walls.  To infer the best navigation strategy to reach a desired objective, the agent employs active inference and utilises the concept of Expected Free Energy (EFE). EFE is a measure of the agent's projected uncertainty or surprise about future states. By minimising EFE, the agent aims to reduce uncertainty and make accurate predictions about future outcomes, thus determining an optimal path to the objective \cite{nav_aif}. This hierarchical active inference process, coupled with the integration of EFE, allows the agent to effectively explore new rooms at the highest level, navigate within the rooms at the mid-level, and execute actions seamlessly at the low level.




The complete generative model is formalised in Appendix~\ref{Appendix_model}, for which we parameterise the likelihood, transition, and approximate posterior mappings of the allocentric and egocentric levels using deep neural networks. These are trained on a dataset of pixel observations collected by sampling random actions in a 3 by 3 rooms mini-grid environment as depicted in Fig\ref{img:3x3_env}, within or across rooms respectively. For more details on the models and training procedure we also refer to Appendix~\ref{app: training_env} and ~\ref{Appendix_model}.

\section{Results}

\subsection{Exploration and goal-oriented search}

We evaluate to what extent the hierarchical active inference model enables our agent to efficiently explore the environment, as well as to then adopt the inferred model for goal-directed planning. Note that our agent is trained task agnostic, and goal-directed behaviour is induced by setting a preferred observation (i.e. the white tile) as typically done in active inference~\cite{AIF_learning}. When the preference is omitted, the agent will be purely driven by epistemic foraging, i.e. maximising information gain, effectively driving exploration.

The exploration task is considered successful when an agent observes 95\% of the visible tiles in the maze, while for the goal-reaching task, the agent must step on the white tile for the task to be considered successful.


We compare our method with four other exploration methods: C-BET \cite{cbet}, Random Network Distillation (RND) \cite{RND}, Curiosity \cite{curiosity}, and Count \cite{count} on both the exploration and goal-seeking task. All baselines are trained in the same conditions as our agent on the same 30 3 by 3 rooms environments.

\begin{table}[b!]
\vspace{-6mm}
\caption{The model's deviation from the mean number of steps (tiles physically covered) provided by the Oracle is evaluated in two tasks within 3 $\times$ 3 rooms environments: full exploration of the maze and reaching the single white tile in the environment.}
\label{tab:steps_exps}

\begin{center}
\begin{small}
\begin{sc}
\begin{tabular}{|l|l|l|l|l|l|}
\hline
task & ours & C-bet & RND & curiosity & count \\ \hline
\begin{tabular}[c]{@{}l@{}}explo-\\ ration\end{tabular} & \textbf{67} & 195 & 177 & 146 & 334 \\ \hline
\begin{tabular}[c]{@{}l@{}}goal \\ reach\end{tabular} & \textbf{34} & 105 & 109 & 121 & 150 \\ \hline
\end{tabular}

\end{sc}
\end{small}

\end{center}

\end{table}

Table \ref{tab:steps_exps} presents the difference in the mean number of steps (number of movements from one tile to the next) between each model and the number of steps needed by the oracle, an A-star algorithm with complete knowledge of the environment layout and agent position. The other models are limited to the information they can observe and must infer the layout and connectivity of the maze. These results indicate that our model outperforms the baselines by requiring fewer steps for both exploration and goal-driven behaviour. Details about the other models' training can be found in Appendix \ref{app: training_env}. Our model reaches objectives faster thanks to its long-term memory capturing the structure of the maze.

\subsection{Qualitative results}

\begin{figure}[b]
\vspace{-5mm}
\begin{center}
\centerline{\includegraphics[width=8cm]{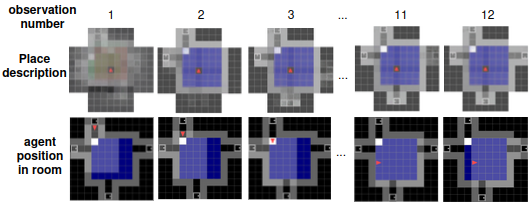}}
\caption{Evolution of the place representation in a room as new observations are provided by the moving agent (red triangle). The model is able to correctly reconstruct the structure of the room as observations are collected.}
\label{img:room_generation}
\end{center}
\vskip -0.1in
\end{figure}

Figure \ref{img:room_generation} illustrates the inference process of place descriptions. Within approximately three steps, the main features of the environment are captured and reasonably accurate based on the accumulated observations. Even when encountering a new aisle for the first time at step 11, the model is able to adapt and generate a well-imagined representation. Each observation corresponds to the red agent's clear field of view, as depicted in the agent position row of the figure.  



More results are added in appendix. For instance, Appendix \ref{Appendix_results} Fig\ref{img:MSE_envs} shows the agent consistently achieving a stable place description within around three observations in room sizes that were part of its training. Interestingly, the agent also exhibits the ability to accurately reconstruct larger rooms, even though it did not encounter such sizes during training. In particular, stable place descriptions for 8-tile wide rooms are attained in approximately five steps. This showcases the agent's allocentric model generalisation abilities beyond the limits of its training.

Moreover, our model is able to disambiguate between visually similar places through the combination of accurate place inference by the allocentric model and the agent location in the cognitive map. This capability is evident in Appendix \ref{Appendix_results} Fig \ref{img:proba_rooms}, where the rooms share a similar appearance (shape or colour). Despite this visual aliasing, the agent successfully distinguishes between them, and a single graph node is created for each room, even when entering from a different door than the initial entry point. This demonstrates the robustness of the agent's perceptual and spatial reasoning abilities in disambiguating similar environments. 

Finally, our hierarchical model also enables accurate predictions over long timescales crossing different rooms, whereas recurrent state space models typically fail when the agent needs to predict across room boundaries. Appendix \ref{Appendix_results} Fig\ref{img:traj_over_layers} depicts each layer's prediction ability over a long-term imagined trajectory. 

\section{Discussion}

Our hierarchical active inference model demonstrates proficiency in learning the maze-like structure of the mini-grid environment and establishing connections between rooms by leveraging the cognitive map's ability to retain spatial locations and their relative relationships. The allocentric model contributes to inferring and learning the structural layout of rooms based on observed positions, while the egocentric model facilitates informed decision-making during actions. In this work, we focused on comparing to other methods that encourage exploration like C-BET \cite{cbet}, Random Network Distillation (RND) \cite{RND}, Curiosity \cite{curiosity} or Count \cite{count}. However, further testing against other models in more diverse environments, including larger environments and bigger rooms is called for, as our model has the capacity to scale to new maze sizes without the need for re-training. In future work, we also intend to benchmark against model-based RL methods~\cite{dreamer}, especially the ones that also learn hierarchical structure in their world model~\cite{choreo}.


One limitation of the current system is that we explicitly train the allocentric model on data collected within one room, which yields proper space representations. To mitigate this, and generalise to arbitrary environments, we could consider splitting the data by unsupervised clustering~\cite{Self-labelling}, or by using the model's prediction error to chunk the data into separate spaces~\cite{verbelen2022chunking}.

While our navigation exhibits strong exploration and goal-reaching abilities, there is still room for improvement. For instance, every time the agent enters a new room, it needs a few steps to relocate in the allocentric reference frame. Adopting a more precise prior on where the agent will enter the room, given the previous room, would further improve this. We could also learn a prior over topological map structures, which would better drive exploration as the agent would expect to discover new rooms, encouraging more information gain at the highest level, and allowing the agent to imagine shortcuts beforehand~\cite{shortcuts}. Additionally, while the current design is pixel-based and provides a top-down view of the agent, applying this model to a 3D first-person view environment, such as the Memory maze \cite{memory_maze}, would bring us closer to real-world applications.

\section*{Acknowledgment}
This research received funding from the Flemish Government under the “Onder-
zoeksprogramma Artificiële Intelligentie (AI) Vlaanderen” programme.


\bibliography{main}
\bibliographystyle{icml2023}

\newpage
\appendix
\onecolumn


\section{Training Environment}
\label{app: training_env}
In this paper all models were trained on a mini-grid environment \cite{gym_minigrid} consisting of 3 by 3 squared rooms of 4 to 7 tiles wide connected by aisles of fixed length randomly placed, separated by a closed door in the middle. Each room is assigned a color at random from a set of four: red, green, blue, and purple. In addition, white tiles may be present at random positions in the map.
The agent's perspective of the environment is represented as a top view, encompassing a window of 7 by 7 tiles. This window includes the tile on which the agent is currently located. However, the agent has no visibility behind itself, and it cannot perceive through walls or closed doors. The observation interpreted by our model is a 3x56x56 RGB pixel rendering, while other models such as  C-BET \cite{cbet}, Random Network Distillation (RND) \cite{RND}, Curiosity \cite{curiosity} and Count \cite{count} utilize a hot-encoded list of 7x7 where each number represents a tile type (wall, floor, goal...). The agent motion in this environment is limited to three actions: going one tile forward or turn 90 degree either left or right.

Both the allocentric and egocentric models underwent independent training using a shared dataset consisting of 400 environments, with 100 environments per room size. The goal of the training process was to efficiently acquire knowledge about the composition of a room and comprehend the dynamic constraints present in the environment, such as walls and doors. The allocentric model was trained on random sequences of varying lengths within a single room, whereas the egocentric model considered the entire random motion spanning 400 steps within the environment. Throughout the training phase, the trajectories followed by the agent were entirely random.


To ensure a fair and rigorous comparison, all other models included in the study were trained and tested on an identical set of environments. These models used their respective policy networks to guide their trajectories, with the objective of maximising performance according to their specific criteria, as specified in \cite{cbet}. By training on the same environments, any differences in performance between the models can be attributed to their distinct approaches and architectures, allowing a meaningful comparison.

\section{Our Model Details}
\label{Appendix_model}

\subsection{Model description}
Results wise we gained in efficiency by keeping the allocentric and egocentric models running and training independently on the same data. Thus in practice, we have the egocentric model learning its latent state through the joint probability of the agent’s observations, actions, policies, belief states and its corresponding approximate posterior. 
It is composed of the transition model $p_\theta$ to incorporate action in its reasoning, the likelihood models $p_\xi$ to construct pixel-based observations, and the posterior model $p_\phi$ to incorporate past events to future state.

\begin{equation} 
P(\tilde{o}, \tilde{c}, \tilde{s}, \tilde{a}) = P(s_{\theta}) \prod_{t=1}^T \overbrace{P(s_{t}|s_{t-1}, a_{t-1})}^{p\theta(s_{t}|s_{t-1}, a_{a_1})} \prod_{t=1}^T \overbrace{P(o_t, c_t|s_t)}^{p_\xi(o_{t} c_t| s_{t})}
\end{equation}

In parallel, the allocentric model forms an internal belief about the world and updates its belief by interacting with it, resulting in a place (latent state z) structured upon positions (p) and corresponding observations (o) \cite{GQN_origin, GQN_Toon} . 
The corresponding true posterior of this allocentric model is:

\begin{equation}
P'(z,\tilde{o}, \tilde{p}) = P(z) \prod_{k=1}^i P(o_k|p_k,z)P(p_k)
\end{equation}

The cognitive map integrates successive action over time steps to obtain the estimated translation and rotation of the agent into a 3D grid using an Attractor Network (CAN) \cite{ratslam_hippo}.
When the current place being defined by the allocentric model differs significantly from any other, a new  experience cell is created.
\FloatBarrier
\subsection{Model Hyperparameters}

\begin{table}[!hbt]
\centering
\scalebox{0.7}{
\begin{tabular}{llcc}
                                               & \textbf{Layer} & \multicolumn{1}{l}{\textbf{Neurons/Filters}} & \multicolumn{1}{l}{\textbf{Stride}} \\ \hline
\multirow{1}{*}{Positional Encoder}             & Linear         & 9                                           &                                     \\ \hline

\multirow{3}{*}{Posterior}                 & Convolutional  & 16                                           & 1 // (kernel:1)                      \\
\multicolumn{1}{c}{}                           & Convolutional  & 32                                           & 2                                   \\
\multicolumn{1}{c}{}                           & Convolutional  & 64                                           & 2                                   \\
\multicolumn{1}{c}{}                           & Convolutional  & 128                                           & 2                                   \\
                                               & Linear         & 2*32                                         &                                     \\ \hline
\multicolumn{1}{c}{\multirow{8}{*}{Likelihood}} & Concatenation  &                                           &                                    \\
                                               & Linear         & 256*4*4                                      &                                     \\
                                               & Upsample       &                                              &                                    \\
                                               & Convolutional  & 128                                          & 1                                   \\
                                               & Upsample       &                                              &                                     \\
                                               & Convolutional  & 64                                           & 1                                   \\
                                               & Upsample       &                                              &                                     \\
                                               & Convolutional  & 32                                           & 1                                   \\
                                               & Upsample       &                                              &                                     \\
                                               & Convolutional  & 3                                            & 1                                   \\ \hline
\multirow{4}{*}{Positional Decoder}            & Concatenation  &                                           &                                    \\
                                               & Linear         & 2048                                          &                                     \\
                                               & Linear         & 512                                      &                                     \\
                                               & Linear         & 128                                          &                                     \\
                                               & Linear         & 64                                          &                                     \\
                                               & Linear         & 9                                         &                                     
                                               \\ \hline
                                            
\end{tabular}
}
\caption{allocentric model parameters}
\label{table:gqn params}
\end{table}

\begin{table}[!hbt]
\centering
\scalebox{0.7}{
\begin{tabular}{llcc}
                                               & \textbf{Layer} & \multicolumn{1}{l}{\textbf{Neurons/Filters}} & \multicolumn{1}{l}{\textbf{Stride}} \\ \hline
\multirow{3}{*}{Prior}                         & Concatenation  &                                              &                                     \\
                                               & LSTM           & 256                                          &                                     \\
                                               & Linear         & 2*32                                         &                                     \\ \hline
\multicolumn{1}{c}{\multirow{8}{*}{Posterior}} & Convolutional  & 8                                           & 2                                   \\
\multicolumn{1}{c}{}                           & Convolutional  & 16                                           & 2                                   \\
\multicolumn{1}{c}{}                           & Convolutional  & 32                                           & 2                                   \\
\multicolumn{1}{c}{}                           & Concatenation  &                                              &                                     \\
\multicolumn{1}{c}{}                           & Linear         & 256                                          &                                     \\
\multicolumn{1}{c}{}                           & Linear         & 64                                         &                                     \\ \hline

\multirow{4}{*}{Image Likelihood}             & Linear         & 256                                          &                                     \\
                                               & Linear         & 32*7*7                                      &                                     \\
                                               & Upsample       &                                              &                                    \\
                                               & Convolutional  & 16                                          & 1                                   \\
                                               & Upsample       &                                              &                                     \\
                                               & Convolutional  & 8                                           & 1                                   \\
                                               & Upsample       &                                              &                                     \\
                                               & Convolutional  & 3                                           & 1                                   \\  \hline
\multirow{3}{*}{Collision Likelihood}         & Linear  &       16                                       &                                     \\
                                               & Linear           & 8                                          &                                     \\
                                               & Linear         & 1                                         &     

                                               \\ \hline
                                            
\end{tabular}
}
\caption{egocentric model parameters}
\label{table:oz params}
\end{table}
\FloatBarrier
\subsection{Navigation details}


During navigation, the cognitive map takes into account the context to avoid looping back to already visited places. If the current understanding of the environment matches a previously encountered place, the corresponding view cell gets activated. However, to avoid ambiguities due to the agent's changing position, the system duplicates the place if it is far from the previous experience, and adapts it to new visual stimuli while preserving the original view cell. In this way, the problem of aliases is effectively resolved.

Moreover, to avoid being trapped in a local minimum while navigating through a large environment with multiple rooms, the agent may encounter a situation where it returns to a known place with no apparent links to unexplored areas. In such cases, the agent is encouraged to visit the least explored or oldest connected place to prevent it from stopping exploration. A decay parameter associated with the view-cell is utilised to achieve this, causing the activity of the current view-cell to decay over time as it is solicited, and all past view-cells to diminish gradually as they are no longer stimulated. This approach ensures that the agent continues to explore new areas and avoids getting stuck in a particular part of the environment.

\section{Supplementary results}
\label{Appendix_results}

\begin{figure}[hbt!]
  \centering
  \begin{minipage}[b]{0.4\textwidth}
    \includegraphics[width=\textwidth]{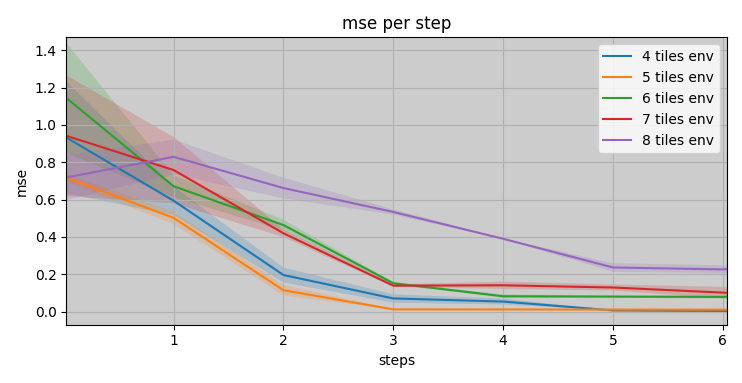}
    \caption{Prediction error of unvisited positions over min 5 tests per 5 environments by room size starting from step 0 where the models has no observation.}
    \label{img:MSE_envs}
  \end{minipage}
  \hfill
  \begin{minipage}[b]{0.5\textwidth}
    \includegraphics[width=\textwidth]{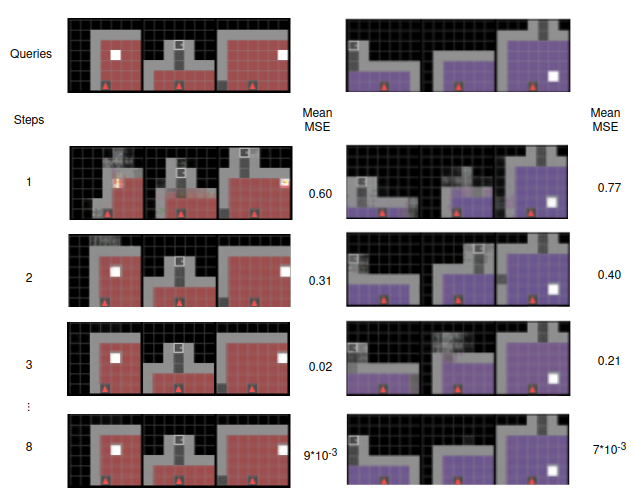}
    \caption{Observation queries, sampled prediction over unvisited position and mean MSE over 5 samples of predictions}
    \label{img:MSE_eg}
  \end{minipage}
\end{figure}

Supplementary tests showed how fast and accurately the allocentric model predicts observations from un-visited positions in a new room. Starting from its place generation and no observation up to 6 observations in several rooms of increasing size, the agent is able to come to a general description of the place in about 3 steps, having an MSE under 0.2 (Fig. \ref{img:MSE_envs}), which is visually close looking as is demonstrated in Fig. \ref{img:MSE_eg}. The agent is also able to solve bigger environments it never encountered in training as 8 tiles wide rooms given more observations (in average 5).

\begin{figure}[hbt!]
\includegraphics[width=16cm]{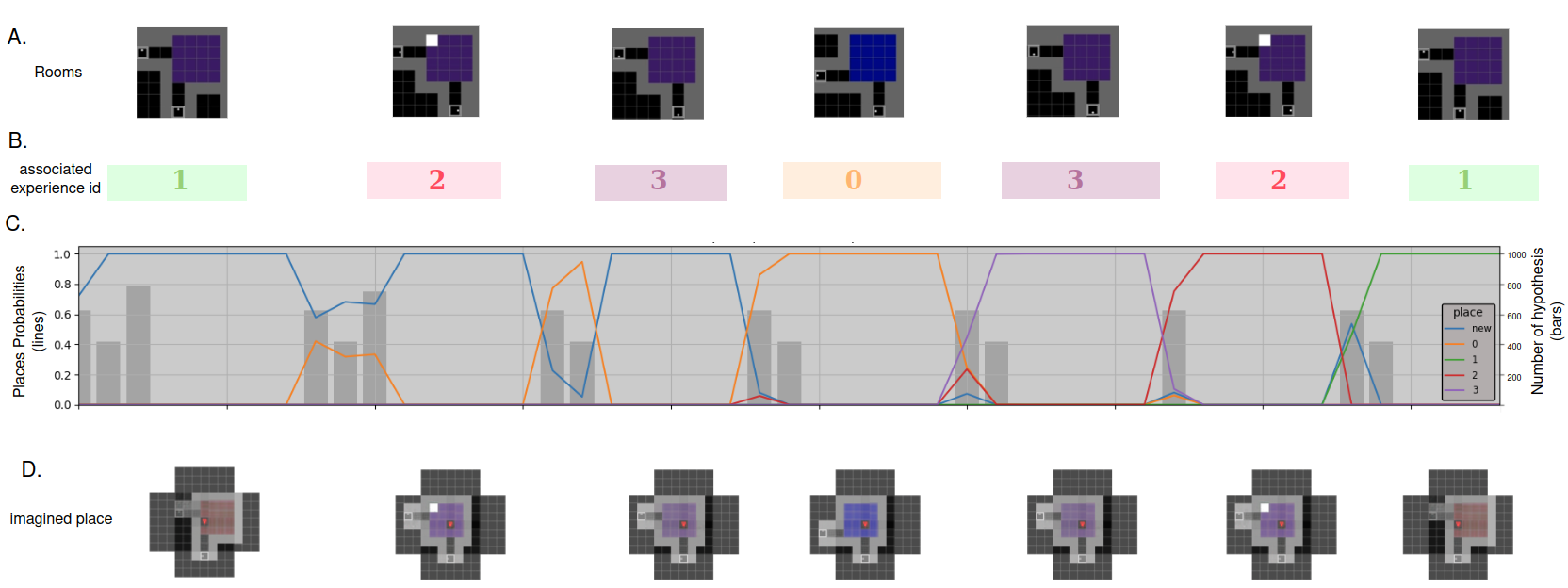}
\caption{Navigation samples of the agent looping clockwise and anti-clockwise (thus entering from a different door) in a new environment of 2 by 2 rooms over 142 steps. The clockwise navigation corresponds to a fully new exploration generating new places (see C.) while the anti-clockwise loop leads through explored places. A.) a new world composed of 4 close looking rooms (colour or/and shapes), B.) the model associated each room to a different experience id corresponding to the place C.) the probability of a new place being created (in blue, the most probable place among all possibilities) or an existing place being deemed the most probable to explain the environment. The grey bars represent how many new places are considered at once, the number of simultaneous hypothesis being considered can be read on the right part of the plot. D.) the imagined place generated for each experience id. We can see that experience 1 is not fully accurate, yet it is enough to distinguish it from the other rooms given real observations of it. }
\label{img:proba_rooms}
\end{figure}

We placed an agent in a new environment consisting of interconnected rooms with similar appearances (in terms of color and/or shape), arranged in a 2x2 pattern, as shown in Figure \ref{img:proba_rooms} A. The agent's task was to move clockwise until it returned to the starting point, and then move anti-clockwise through the previously visited rooms.

Whenever the agent identified a new location, it created a new view-cell and experience ID to store the information, as depicted in Figure \ref{img:proba_rooms} B. If the prediction error increased above a threshold of 0.5 MSE, indicating that the previous place description didn't match the current observation, the agent would rely on the previously known view-cells, based on their probability, to describe the current room, as shown in Figure \ref{img:proba_rooms} C. The bars in the figure represent the amount of hypotheses being considered simultaneously to explain the agent's observations. We can see that upon entering a new place, the agent started investigating around 600 different hypotheses to determine its current position, with a third of the worst hypotheses being erased at each step. As the agent's best hypothesis converged to an understanding of the place, some new hypotheses were also considered. The blue line in Figure \ref{img:proba_rooms} C. represents the probability of the best new hypothesis compared to the known places' hypotheses. The agent usually converged to a precise understanding of a new place within four steps, whereas it took about two steps in an already known room, thanks to the probability of the previous place outweighing any new place description, which matched observations more precisely.

Figure \ref{img:proba_rooms} D. depicts the imagined rooms used for hypothesis creation when entering a new room, with all door entries being considered to create new hypotheses, which were immediately discarded if they did not match observations. The associated probability of the view-cell was proportional to the precision of the predictions, and a single view-cell could have several hypotheses running at once. In the figure, only the best hypothesis was considered. The imagined rooms were almost correctly inferred, with experience 1 having a less precise reconstruction, although it was still sufficient to recognize it when entering from a new door.

\begin{figure}[hbt!]
\includegraphics[width=16cm]{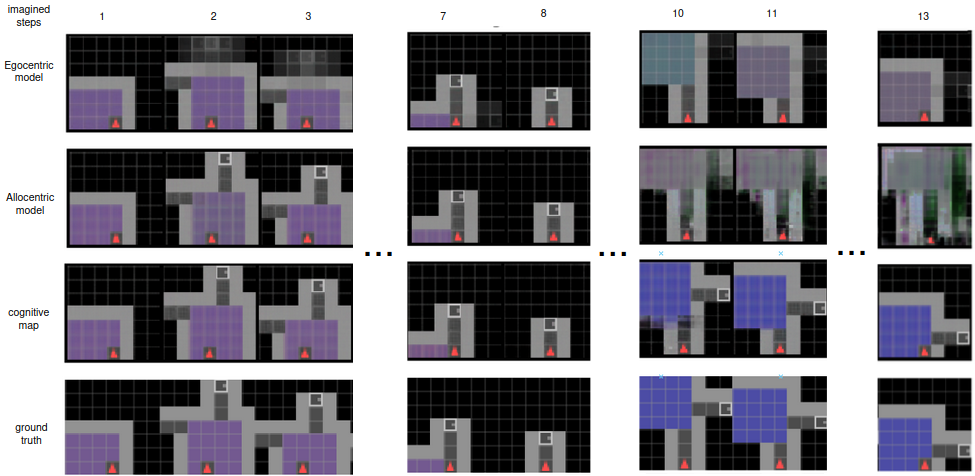}
\caption{An imagined trajectory by each level of a path leading toward an already visited room .
The egocentric model has a short-term memory therefore it forgets information as time passes, as can be observed starting step 2, the front aisle is already missing after the agent turned a few times without seeing it. The allocentric model however doesn't forget the place description over time but gets lost once leaving the place it is currently in. While the cognitive map knows the link between locations and is able to correctly infers the place to be expected behind the door giving a result much similar to the ground truth.}
\label{img:traj_over_layers}
\end{figure}


\end{document}